\documentclass[11pt,letterpaper]{article}
\usepackage[dvipsnames]{xcolor}
\usepackage[hyperref]{ijcnlp2017}
\usepackage{times}
\usepackage{latexsym}
\usepackage{url}
\usepackage{stmaryrd}
\usepackage{amsmath}
\usepackage{booktabs}
\usepackage{color}
\usepackage[colorinlistoftodos,prependcaption,textsize=small]{todonotes}
\usepackage{multirow}
\usepackage{subcaption}
\usepackage{graphicx}
\usepackage{amssymb}
\usepackage{xspace}
\usepackage{siunitx}

\newcommand \flickr{\textsc{Flickr30K}\xspace}
\usepackage{pifont}
\ijcnlpfinalcopy

\title{Natural Language Inference from Multiple Premises}

\author{
Alice Lai$^1$\ \ \ \ \ \ \ \ \ \ Yonatan Bisk$^2$\ \ \ \ \ \ \ \ \ \ Julia Hockenmaier$^1$\\
  $^1$Department of Computer Science, University of Illinois at Urbana-Champaign\\
  $^2$Paul G. Allen School of Computer Science \& Engineering, Univ. of Washington\\
  {\normalsize \href{mailto:aylai2@illinois.edu}{\nolinkurl{aylai2@illinois.edu}}, 
  \href{mailto:ybisk@cs.washington.edu}{\nolinkurl{ybisk@cs.washington.edu}},
  \href{mailto:juliahmr@illinois.edu}{\nolinkurl{juliahmr@illinois.edu}}}
  }

\date{}

\begin{document}

\maketitle

\begin{abstract}
  We define a novel textual entailment task that requires inference over multiple premise sentences. We present a new dataset for this task that minimizes trivial lexical inferences, emphasizes knowledge of everyday events, and presents a more challenging setting for textual entailment. We evaluate several strong neural baselines and analyze how the multiple premise task differs from standard textual entailment.
\end{abstract}

\section{Introduction}
Standard textual entailment recognition is concerned with deciding whether one statement (the hypothesis) follows from another statement (the premise). 
However, in some situations, multiple independent descriptions of the same event are available, e.g. multiple news articles describing the same story, social media posts by different people about a single event, or multiple witness reports for a crime. In these cases, we want to use multiple independent reports to infer what really happened. 

We therefore introduce a variant of the standard textual entailment task in which the premise text consists of multiple independently written sentences, all describing the same scene (see examples in Figure~\ref{fig:mpe_example}).  The task is to decide whether the hypothesis sentence 1) can be used to describe the same scene (entailment), 2) cannot be used to describe the same scene (contradiction), or 3) may or may not describe the same scene (neutral). The main challenge is to infer what happened in the scene from the multiple premise statements, in some cases aggregating information across multiple sentences into a coherent whole.

Similar to the SICK and SNLI datasets \cite{sick2014,Bowman2015}, each premise sentence in our data is a single sentence describing everyday events, rather than news paragraphs as in the RTE datasets \cite{Dagan:2005}, which require named entity recognition and coreference resolution. Instead of soliciting humans to write new hypotheses, as SNLI did, we use simplified versions of existing image captions, and use a word overlap filter and the structure of the denotation graph of \newcite{young2014} to minimize the presence of trivial lexical relationships.

\begin{figure}
\begin{small}
\centering
\begin{tabular}{@{}p{7.5cm}@{}}
\textbf{Premises:}\\
1. Two girls sitting down and looking at a book.\\
2. A couple laughs together as they read a book on a train.   \\ 
3. Two travelers on a train or bus reading a book together. \\ 
4. A woman wearing glasses and a brown beanie next to      \\ 
\hspace{9pt}   a girl with long brown hair holding a book. \\
\textbf{Hypothesis:}\\
\hspace{9pt}Women smiling.   \hfill $\Rightarrow$\textsc{\textbf{Entailment}}\\[5pt]
\textbf{Premises:}\\
1. Three men are working construction on top of a building.\\
2. Three male construction workers on a roof working \\
\hspace{9pt} in the sun.   \\ 
3. One man is shirtless while the other two men work \\
\hspace{9pt} on construction. \\ 
4. Two construction workers working on infrastructure, \\
\hspace{9pt} while one worker takes a break.      \\ 
\textbf{Hypothesis:} \\
\hspace{9pt}A man smoking a cigarette.    \hfill $\Rightarrow$\textsc{\textbf{Neutral}}\\[5pt]
\textbf{Premises:}\\
1. A group of individuals performed in front of a seated \\
\hspace{9pt} crowd.\\
2. Woman standing in front of group with black folders in \\
\hspace{9pt} hand.   \\ 
3. A group of women with black binders stand in front of a\\
\hspace{9pt} group of people.  \\ 
4. A group of people are standing at the front of the room,\\ 
\hspace{9pt} preparing to sing. \\ 
\textbf{Hypothesis:} \\
\hspace{9pt} A group having a meeting.    \hfill $\Rightarrow$\textsc{\textbf{Contradiction}}
\end{tabular}
\end{small}
\vspace{-6pt}
\caption{The Multiple Premise Entailment Task}
\vspace{-13pt}
\label{fig:mpe_example}
\end{figure}


\section{Related Standard  Entailment Tasks}

In the following datasets, premises are single sentences drawn from image or video caption data that describe concrete, everyday activities.

The \textbf{SICK} dataset \cite{sick2014} consists of 10K sentence pairs. The premise sentences come from the \textsc{Flickr8K} image caption corpus \cite{rashtchian2010collecting} and the MSR Video Paraphrase Corpus \cite{agirre2012semeval}, while the hypotheses were automatically generated. This process introduced some errors (e.g. ``A motorcycle is riding standing up on the seat of the vehicle'') and an uneven distribution of phenomena across entailment classes that is easy to exploit (e.g. negation \cite{lai2014semeval}).

The \textbf{SNLI} dataset \cite{Bowman2015} contains over 570K  sentence pairs. The premises come from the \flickr image caption corpus \cite{young2014} and VisualGenome \cite{krishnavisualgenome}. The hypotheses were written by Mechanical Turk workers who were given the premise and asked to write one definitely true sentence, one possibly true sentence, and one definitely false sentence. The task design prompted workers to write hypotheses that frequently parallel the premise in structure and vocabulary, and therefore the semantic relationships between premise and hypothesis are often limited to synonym/hyponym lexical substitution, replacement of short phrases, or exact word matching.

\section{The Multiple Premise Entailment Task}\label{mpe}
In this paper, we propose a variant of entailment where each hypothesis sentence is paired with an unordered set of independently written premise sentences that describe the same event. The premises may contain overlapping information, but are typically not paraphrases. The majority of our dataset requires consideration of multiple premises, including aggregation of information from multiple sentences.

This Multiple Premise Entailment (MPE) task is inspired by the Approximate Textual Entailment (ATE) task of \newcite{young2014}. Each item in the ATE dataset consists of a premise set of four captions from \flickr, and a short phrase as the hypothesis. The ATE data was created automatically, under the assumption that items are positive (approximately entailing) if the hypothesis comes from the same image as the four premises, and negative otherwise. However, Young~et~al. found that this assumption was only true for just over half of the positive items. For MPE, we also start with four \flickr captions as the premises and a related/unrelated sentence as the hypothesis, but we restrict the hypothesis to have low word overlap with the premises, and we collect human judgments to label the items as entailing, contradictory, or neutral.

\section{The MPE Dataset}
\label{sec:data}
The MPE dataset (Figure~\ref{fig:mpe_example}) contains 10,000 items (8,000 training, 1,000 development and 1,000 test), each consisting of four premise sentences (captions from the same \flickr image), one hypothesis sentence (a simplified \flickr caption), and one label (entailment, neutral, or contradiction) that indicates the relationship between the set of four premises and the hypothesis. This label is based on a consensus of five crowdsourced judgments. To analyze the difference between multiple premise and single premise entailment (Section \ref{sec:multi_premise}), we also collected pair label annotations for each individual premise-hypothesis pair in the development data. This section describes how we selected the premise and hypothesis sentences, and how we labeled the items via crowdsourcing.

\subsection{Generating the Items}

\paragraph{Hypothesis simplification}
The four premise sentences of each MPE item consist of four original \flickr captions from the same image. Since complete captions are too specific and are likely to introduce new details that are not entailed by the premises, the hypotheses sentences are simplified versions of \flickr captions. Each hypothesis sentence is either a simplified variant of the fifth caption of the same image as the premises, or a simplified variant of one of the captions of a random, unrelated image. 

Our simplification process relies on the denotation graph of \newcite{young2014}, a subsumption hierarchy over phrases, constructed from the captions in \flickr. They define a set of normalization and reduction rules (e.g. lemmatization, dropping modifiers and prepositional phrases, replacing nouns with their hypernyms, extracting noun phrases) to transform the original captions into shorter, more generic phrases that are still true descriptions of the original image. 

To simplify a hypothesis caption, we consider all sentence nodes in the denotation graph that are ancestors (more generic versions) of this caption, but exclude nodes that are also ancestors of any of the premises. 
This ensures that the simplified hypothesis cannot be  trivially obtained from a premise via the same automatic simplification procedure. Therefore, we avoid some obvious semantic relationships between premises and hypothesis, such as hypernym replacement, dropping modifiers or PPs, etc.

\paragraph{Limiting lexical overlap}
Given the set of simplified, restricted hypotheses, we further restrict the pool of potential items to contain only pairings where the hypothesis has a word overlap $\leq$ 0.5 with the premise set. We compute word overlap as the fraction of hypothesis tokens that  appear in  at least one  premise (after stopword removal).  This eliminates trivial cases of entailment where the hypothesis is simply a subset of the premise text.
 Table~\ref{tab:word_overlap} shows that the mean word overlap for our training data is much lower than SNLI. 

\begin{table}[ht]
\begin{center}
\begin{small}
\begin{tabular}{@{}l@{\hspace{4pt}}r@{\hspace{4pt}}l@{\hspace{8pt}}r@{\hspace{4pt}}l@{\hspace{8pt}}r@{\hspace{4pt}}l@{\hspace{8pt}}r@{\hspace{4pt}}l@{}}
	\toprule
     & \multicolumn{4}{c}{SNLI} & \multicolumn{4}{c}{MPE} \\
    Data & \multicolumn{2}{c}{full} & \multicolumn{2}{c}{lemma } & \multicolumn{2}{c}{full} & \multicolumn{2}{c}{lemma } \\
    \midrule
	All & 0.44 & {\scriptsize $\pm$ 0.29 } & 0.48 & {\scriptsize $\pm$ 0.29 } & 0.28 & {\scriptsize $\pm$ 0.22} & 0.33 & {\scriptsize $\pm$ 0.20 } \\
    E & 0.59 & {\scriptsize $\pm$ 0.31 } & 0.64 & {\scriptsize $\pm$ 0.30 } & 0.34 & {\scriptsize $\pm$ 0.21} & 0.38 & {\scriptsize $\pm$ 0.19 } \\
    N & 0.41 & {\scriptsize $\pm$ 0.24 } & 0.45 & {\scriptsize $\pm$ 0.24 } & 0.28 & {\scriptsize $\pm$ 0.21} & 0.33 & {\scriptsize $\pm$ 0.19 }  \\
    C & 0.33 & {\scriptsize $\pm$ 0.25 } & 0.36 & {\scriptsize $\pm$ 0.25 } & 0.23 & {\scriptsize $\pm$ 0.22} & 0.30 & {\scriptsize $\pm$ 0.21 } \\
	\bottomrule
\end{tabular}
\end{small}
\caption{Mean word overlap for full training data and each label, original and lemmatized sentences. MPE has much lower word overlap than SNLI.}
\vspace{-10pt}
\label{tab:word_overlap}
\end{center}
\end{table}

\paragraph{Data selection}
From this constrained pool of premises-hypothesis pairings, we randomly sampled 8000 items from the \flickr training split for our training data. For test and development data, we sample 1000 items from \flickr test and 1000 from dev. The hypotheses in the training data must be associated with at least two captions in the \flickr train split, while the hypotheses in dev/test must be associated with at least two captions in the union of the training and dev/test, and with at least one caption in dev/test alone. Since the test and dev splits of \flickr are smaller than the training split, this threshold selects hypotheses that are rare enough to be interesting and frequent enough to be reasonable sentences.

\begin{table*}
\centering
\begin{small}
\begin{tabular}{@{}p{16cm}@{}}	
\textbf{Instructions:}\\
We will show you four caption sentences that describe the same scene, and one proposed sentence. Your task is to decide whether or not the scene described by the four captions can also be described by the proposed sentence.\\
The four captions were written by four different people. All four people were shown the same image, and then wrote a sentence describing the scene in this image. Therefore, there may be slight disagreements among the captions. The images are photographs from Flickr that show everyday scenes, activities, and events. You will not be given the image that the caption writers saw.\\[4pt] 

\textbf{Process:}\\
Read the four caption sentences and then read the proposed sentence.\\
Choose 1 of 3 possible responses to the question \\
\textbf{Can the scene described by the four captions also be described by the proposed sentence?}\\[4pt]

 \texttt{\textbf{Yes}}: \hspace{4pt}
 The scene described by the captions can definitely (or very probably) be described by the proposed sentence.
The proposed sentence may leave out details that are mentioned in the captions. If the proposed sentence describes something that is not mentioned in the captions, it is probably safe to assume the extra information is true, given what you know from the captions. If there are disagreements among the captions about the details of the scene, the proposed sentence is consistent with at least one caption.\\

 \texttt{\textbf{Unknown}}:\hspace{4pt} There is not enough information to decide whether or not the scene described by the captions can be described by the proposed sentence.
There may be scenes that can be described by the proposed sentence and the captions, but you don't know whether this is the case here.\\

 \texttt{\textbf{No}}:\hspace{4pt} The scene described by the captions can probably not be described by the proposed sentence.
The proposed sentence and the captions either contradict each other or describe what appear to be two completely separate events.\\

\end{tabular}
\end{small}
\caption{The annotation instructions we provided to Crowdflower and Mechanical Turk annotators.}
\label{tab:turk_instructions}
\end{table*}

\subsection{Assigning Entailment Labels}
\label{sec:label_crowdsource}
\paragraph{Crowdsourcing procedure} 
For each item, we solicited five responses from Crowdflower and Amazon Mechanical Turk as to whether the hypothesis was \textit{entailed}, \textit{contradictory}, or \textit{neither} given a set of four premises. Instructions are shown in Table~\ref{tab:turk_instructions}. We provided labeled examples to illustrate the kinds of assumptions we expected. 

\paragraph{Entailment labels} We assume three labels (entailment, neutral, contradiction). For entailment, we deliberately asked annotators to judge whether the hypothesis could \textit{very probably} describe the same scene as the premises, rather than specifying that the hypothesis must \textit{definitely} be true, as \newcite{Bowman2015} did for SNLI. Our instructions  align with the standard definition of textual entailment: ``T entails H if humans reading T would typically infer that H is most likely true'' \cite{dagan2013book}. We are not only interested in what is logically required for a hypothesis to be true, but also in what human readers assume is true, given their own world knowledge. 


\paragraph{Final label assignment}
Of the 10,000 items for which we collected full label annotations, 90\% had a majority label based on the five judgments, including 16\% with a 3-2 split between entailment and contradiction. The remaining 10\% had a 2-2-1 split across the three classes. We manually adjudicated the latter two cases. As a result, 82\% of the final labels in the dataset correspond to a majority vote over the judgments (the remaining 18\% differ due to our manual correction). The released dataset contains both our final labels and the crowdsourced judgments for all items.

\paragraph{Image IDs}
Premises in the our dataset have corresponding image IDs from \flickr. 
We are interested in the information present in linguistic descriptions of a scene, so our labels reflect the textual entailment relationship between the premise text and the hypothesis. Future work could apply multi-modal representations to this task, with the caveat that the image would likely resolve many neutral items to either entailment or contradiction.


\section{Data Analysis}
\subsection{Statistics}
The dataset contains 8000 training items, 1000 development items, and 1000 test items. Table~\ref{tab:train_stats} shows overall type and token counts and sentence lengths as well as the label distribution.
\begin{table}[ht]
\begin{center}
\begin{small}
\begin{tabular}{lrr}
	\toprule
     & \textbf{SNLI} & \textbf{MPE} \\
	\#Lexical types & 36,616 & 9,254 \\
    \#Lexical tokens & 12 million & 468,524 \\
    Mean premise length & 14.0 {\scriptsize $\pm$ 6.0} & 53.2 {\scriptsize $\pm$ 12.8} \\
    Mean hypothesis length & 8.3 {\scriptsize $\pm$ 3.2} & 5.3 {\scriptsize $\pm$ 1.8}  \\
    \midrule
    \textbf{Label distribution} & \\
    Entailment & 33.3\% & 32.3\%  \\
    Neutral & 33.3\% & 26.3\%  \\
    Contradiction & 33.3\% & 41.6\%  \\
	\bottomrule
\end{tabular}
\end{small}
\caption{Type and token counts, sentence lengths, and label distributions for training data.}
\vspace{-10pt}
\label{tab:train_stats}
\end{center}
\end{table}

The mean annotator agreement, i.e. the fraction of annotators who agreed with the final label, is 0.70 for the full dataset, or 0.82 for the entailment class, 0.42 for neutral, and 0.78 for contradiction. That is, on average, four of the five crowdsourced judgments agree with the final label for the entailment and contradiction items, whereas for the neutral items, only an average of two of the five original annotators assigned the neutral label, and the other three were split between contradiction and entailment.

\begin{table*}[ht]
\begin{center}
\begin{footnotesize}
\begin{tabular}{p{0.9cm}@{\hspace{6pt}}l@{\hspace{6pt}}p{1cm}@{\hspace{4pt}}p{11.5cm}@{}}
	\toprule
    \# pairs agree & \% of data & Pair Label & Example \textbf{Hypothesis} and Four Premises \\
    \midrule
0 & 21.8 & N \newline N \newline N \newline N  & A football player in a red uniform is standing in front of other football players in a stadium. \newline
	A football player facing off against two others.\newline
	A football player wearing a red shirt. \newline
	Defensive player waiting for the snap. \newline
    $\Rightarrow$\textsc{\textbf{E}} \hspace{4pt} \textbf{The team waiting.}  
\\[4pt] 
  1 & 26.9 & N \newline C \newline N \newline N \newline & 
    A person is half submerged in water in their yellow kayak.\newline
    A woman has positioned her kayak nose down in the water.\newline
    A person in a canoe is rafting in wild waters. \newline
    A kayaker plunges into the river.  \newline
    $\Rightarrow$\textsc{\textbf{C}} \hspace{4pt} \textbf{A man in a boat paddling through waters.}
\\[4pt] 
    2 & 16.7 & E \newline E \newline N \newline N \newline & 
    	A batter playing cricket missed the ball and the person behind him is catching it.\newline 		
    	A cricket player misses the pitch.\newline
        The three men are playing cricket. \newline
        A man struck out playing cricket. \newline
       $\Rightarrow$\textsc{\textbf{E}} \hspace{4pt} \textbf{A man swings a bat.}  
\\[4pt] 
  3 & 24.8 & N \newline N \newline E \newline N \newline
    & A young gymnast, jumps high in the air, while performing on a balance beam. \newline 
    	A gymnast performing on the balance beam in front of an audience. \newline 
        The young gymnast's supple body soars above the balance beam. \newline 
        A gymnast is performing on the balance beam. \newline 
        $\Rightarrow$\textsc{\textbf{N}}  \hspace{4pt} \textbf{A woman doing gymnastics.}  
\\[4pt]
    4 & 9.8 & C \newline C \newline C \newline C \newline &
 		A man with a cowboy hat is riding a horse that is jumping.\newline
      A cowboy riding on his horse that is jumping in the air.\newline
      A cowboy balances on his horse in a rodeo.\newline
      Man wearing a cowboy hat riding a horse.\newline
      $\Rightarrow$\textsc{\textbf{C}} \hspace{4pt} \textbf{Men pulled by animals.}   \\
\bottomrule
\end{tabular}
\end{footnotesize}
\caption{MPE examples that illustrate the difference between pair labels and the full label. We include one example for each category, based on the number of pair labels that agree with the full label, and indicate the size of each category as a percentage of the development data.}
\label{tab:examples_difficulty}
\end{center}
\end{table*}

\begin{table*}[ht]
\begin{center}
\begin{footnotesize}
\begin{tabular}{@{}p{1.7cm}@{\hspace{12pt}}r@{\hspace{8pt}}r@{\hspace{8pt}}r@{\hspace{8pt}}r@{\hspace{12pt}}p{11cm}@{}}
	\toprule
     & \# & E & N & C & Example Premise and \textbf{Hypothesis} Pair \\
    \midrule
    Total & 100 & 31 & 29 & 40 & \\
    \midrule
	Word equivalence & 16 & 12 & 4 & 0 & 
	A person \textit{climbing} a rock face.\newline
	\textbf{A rock climber \textit{scales} a cliff.}  $\Rightarrow$\textsc{\textbf{E}}\\[4pt]
    Word hypernymy & 19 & 6 & 6 & 7 & 
    \textit{Girl} in a blue sweater painting while looking at a bird in a book.\newline
 	\textbf{A \textit{child} painting a picture.}  $\Rightarrow$\textsc{\textbf{E}}\\[4pt]
    Phrase equivalence & 7 & 6 & 1 & 0 & 
    \textit{A couple in their wedding attire} stand behind a table with a wedding cake and flowers.\newline
	\textbf{\textit{Newlyweds} standing.}  $\Rightarrow$\textsc{\textbf{E}}\\[4pt]
	Phrase hypernymy & 8 & 6 & 2 & 0 & 
	A group of young boys wearing track jackets \textit{stretch their legs} on a gym floor as they sit in a circle.\newline
	\textbf{A group \textit{doing exercises}.}  $\Rightarrow$\textsc{\textbf{E}}\\[4pt]
    Mutual exclusion & 25 & 0 & 0 & 25 &
	A woman in a red vest \textit{working at a computer}.\newline
 	\textbf{Lady \textit{doing yoga}.}  $\Rightarrow$\textsc{\textbf{C}}\\[4pt]
	Compatibility & 18 & 0 & 18 & 0 &
    \textit{Onlookers watch}.\newline
    \textbf{A girl at bat in a \textit{softball game}.}  $\Rightarrow$\textsc{\textbf{N}} \\[4pt]
    World knowledge & 35 & 14 & 9 & 12 &
	A young woman gives directions to an older woman outside a subway station.\newline
    \textbf{Women standing.} $\Rightarrow$\textsc{\textbf{E}} \\
\bottomrule
\end{tabular}
\end{footnotesize}
\caption{Analysis of 100 random dev items. For each phenomenon, we show the distribution over labels and an example. The label is indicated with E, N, C. We use color and underlining to indicate the relevant comparisons. The indicated span of text is part of the necessary information to predict the correct label, but may not be sufficient on its own.}
\vspace{-10pt}
\label{tab:info_type_pairs}
\end{center}
\end{table*}

\subsection{MPE vs. Standard Entailment}
\label{sec:multi_premise}

Multiple premise entailment (MPE) differs from standard single premise entailment (SPE) in that each premise consists of four independently written sentences about the same scene.  To understand how  MPE differs from SPE, we used crowdsourcing to collect pairwise single-premise entailment labels for each individual premise-hypothesis pair in the development data. Each consensus label is based on three judgments. 

In Table~\ref{tab:examples_difficulty}, we compare the full MPE entailment labels (bold $\Rightarrow$\textsc{\textbf{E}}, $\Rightarrow$\textsc{\textbf{N}}, $\Rightarrow$\textsc{\textbf{C}}), to the four pair SPE labels (E, N, C). The number of SPE labels that agree with the MPE label yields the five categories in Table~\ref{tab:examples_difficulty}, ranging from the most difficult case where none of the SPE labels agree with the MPE label (21.8\% of the data) to the simplest case where all four SPE labels agree with the MPE label (9.8\% of the data).  

We observe that a simple majority voting scheme over the gold standard SPE labels would not be sufficient, since it assigns the correct MPE label to only 34.6\% of the development items (i.e. those cases where three or four SPE pairs agree with the MPE label). 
We also evaluate a slightly more sophisticated voting scheme that applies the following heuristic (here, $E$, $N$, $C$ are the number of SPE labels of each class): 
\begin{center}
\begin{tabular}{l}
If $E > C$, predict entailment.\\
Else if $C > E$, predict contradiction.\\
Otherwise, predict neutral.
\end{tabular}
\end{center}
This baseline achieves an accuracy of  41.7\%. 
These results indicate that MPE cannot be trivially reduced to SPE. That is, even if a model had access to the correct SPE label for each individual premise (an unrealistic assumption), it would require more than simple voting heuristics to obtain the correct MPE label from these pairwise labels. Table~\ref{tab:examples_difficulty} illustrates that the majority of MPE items require aggregation of information about the described entities and events across multiple premises. In the first example, the first premise is consistent with a scene that involves a team of football players, while only the last premise indicates that the team may be waiting.
Moreover, the simple majority voting would work on the fourth example but fail on the second example, while the more sophisticated voting scheme would work on the second example and fail on the fourth.

\subsection{Semantic Phenomena}
\label{sec:discussion}
We used a random sample of 100 development items to examine the types of semantic phenomena that are useful for inference in this dataset. We categorized each item by type of knowledge or reasoning necessary to predict the correct label for the hypothesis given the premises. An item belongs to a category if at least one premise in that item exhibits that semantic phenomenon in relation to the hypothesis, and an item may belong to multiple categories. For each category, Table \ref{tab:info_type_pairs} contains its frequency, an illustrative example containing the relevant premise, and the distribution over entailment labels. We did our analysis on full items (four premises and the corresponding hypothesis), but the examples in Table~\ref{tab:info_type_pairs} have been simplified to a single premise for simplicity.

\paragraph{Word equivalence}\label{word_equiv}
Items in this category contain a pair of equivalent words (synonyms or paraphrases). The word in the hypothesis can be exchanged for the word in the premise without significantly changing the meaning of the hypothesis.

\paragraph{Word hypernymy}\label{hypernym}
These items involve lexical hypernyms: someone who is a man is also a person (entailment), but a person may or may not be a man (neutral), and somebody who is a man is not a child (contradiction).

\paragraph{Phrase equivalence}\label{phrase_equiv}
These items involve equivalent phrases, i.e. synonyms or paraphrases. The phrase in the hypothesis can be replaced by the phrase in the premise without significantly changing the meaning of the hypothesis. 

\paragraph{Phrase hypernymy}\label{phrase_hypernym}
Items in this category involve a specific phrase and a general phrase: the more general phrase ``doing exercises'' can refer to multiple types of exercises in addition to ``stretching their legs.''

\paragraph{Mutual exclusion}\label{mutex}
Distinguishing between contradiction and neutral items involves identifying actions that are mutually exclusive, i.e. cannot be performed simultaneously by the same agent (``Two doctors perform surgery'' vs.  ``Two surgeons are having lunch''). 

\paragraph{Compatibility}\label{compatible}
The opposite of mutual exclusion is compatibility: two actions that can be performed simultaneously by the same agent (e.g. ``A boy flying a red and white kite'' vs. ``A boy is smiling'').

\paragraph{World knowledge}\label{world_knowledge}
These items require extra-linguistic knowledge about the relative frequency and co-occurrence of events in the world (not overlapping with the mutual exclusion or compatibility phenomena). A human reader can infer that children in a potato sack race are having fun (while a marathon runner competing in a race might not be described as having fun). 

\subsection{Combining Information Across Premises}

In addition to the semantic phenomena we have just discussed, the data presents the challenge of how to combine information across multiple premises. We examined examples from the development data to analyze the different types of information aggregation present in our dataset. 

\paragraph{Coreference resolution} This case requires cross-caption coreference resolution of entity mentions from multiple premises and the hypothesis. In this example, a human reader can recognize that ``two men'' and ``two senior citizens'' refer to the same entities, i.e. the ``two older men'' in the hypothesis. Given that information, the reader can additionally infer that the two older men on the street are likely to be standing. 

\begin{center}
\begin{footnotesize}
\begin{tabular}{@{}p{\linewidth}@{}}
1. Two men in tan coats exchange looks on the city sidewalk.	\\
2. Two senior citizens talking on a public street.	 \\ 
3. Two men in brown coats on the street. \\ 
4. Two men in beige coats, talking.	   \\
\midrule
Two older men stand.	
   \hfill $\Rightarrow$\textsc{\textbf{Entailment}}\\
\end{tabular}
\end{footnotesize}
\end{center}

\paragraph{Event resolution} This case requires resolving various event descriptions from multiple premises and the hypothesis. In the following example, a human reader recognizes that the man is sitting on scaffolding so that he can repair the building, and therefore he is doing construction work.

\begin{center}
\begin{footnotesize}
\begin{tabular}{@{}p{\linewidth}@{}}
1. A man is sitting on a scaffolding in front a white building.	\\
2. A man is sitting on a platform next to a building ledge.	\\
3. A man looks down from his balcony from a stone building.	\\
4. Repairing the front of an old building.	\\
\midrule
A man doing construction work.	 \hfill $\Rightarrow$\textsc{\textbf{Entailment}}\\
\end{tabular}
\end{footnotesize}
\end{center}

\paragraph{Visual ambiguity resolution}

This case involves reconciling apparently contradictory information across premises. These discrepancies are largely due to the fact that the premise captions were written to describe an image. Sometimes the image contained visually ambiguous entities or events that are then described by different caption writers. In this example, in order to resolve the discrepancy, the reader must recognize from context that ``woman'' and ``young child'' (also ``person'') refer to the same entity.

\begin{center}
\begin{footnotesize}
\begin{tabular}{@{}p{\linewidth}@{}}
1. A person in a green jacket and pants appears to be digging in a wooded field with several cars in the background.	\\
2.A young child in a green jacket rakes leaves.	\\
3. A young child rakes leaves in a wooded area.	\\
4. A woman cleaning up a park.	\\
\midrule
A woman standing in the forest.	
\hfill $\Rightarrow$\textsc{\textbf{Entailment}}\\
\end{tabular}
\end{footnotesize}
\end{center}

\paragraph{Scene resolution}

These examples require the reader to build a mental representation of the scene from the premises in order to assess the probability that the hypothesis is true. In the first example, specific descriptions -- a jumping horse, a cowboy balancing, a rodeo -- combine to assign a high probability that the specific event described by the hypothesis is true.



\begin{center}
\begin{footnotesize}
\begin{tabular}{@{}p{\linewidth}@{}}
1. A man with a cowboy hat is riding a horse that is jumping.	\\
2. A cowboy riding on his horse that is jumping in the air.	\\
3. A cowboy balances on his horse in a rodeo.	\\
4. Man wearing a cowboy hat riding a horse.	\\
\midrule
An animal bucking a man.\hfill $\Rightarrow$\textsc{\textbf{Entailment}}\\
\end{tabular}
\end{footnotesize}
\end{center}

In the next example, the hypothesis does not contradict any individual premise sentence. However, a reader who understands the generic scene described knows that the very specific hypothesis description is unlikely to go unmentioned. Shirtlessness would be a salient detail in the this scene, so the fact that none of the premises mention it means that the hypothesis is likely to be false. 





\begin{center}
\begin{footnotesize}
\begin{tabular}{@{}p{\linewidth}@{}}
1. A young couple sits in a park eating ice cream as children play and other people enjoy themselves around them.	\\
2. Couple in park eating ice cream cones with three other adults and two children in background.	\\
3. A couple enjoying ice cream outside on a nice day.\\
4. A couple eats ice cream in the park.	\\
\midrule
A shirtless man sitting.	\hfill $\Rightarrow$\textsc{\textbf{Contradiction}}\\
\end{tabular}
\end{footnotesize}
\end{center}






In the final example, the premises present a somewhat generic description of the scene. While some premises lean towards entailment (a woman and a man in \textit{discussion} could be having a work meeting) and others lean towards contradiction (two people conversing outdoors at a restaurant are probably not working), none of them contain overwhelming evidence that the scene entails or contradicts the hypothesis. Therefore, the hypothesis is neutral given the premises.

\begin{center}
\begin{footnotesize}
\begin{tabular}{@{}p{\linewidth}@{}}
1. A blond woman wearing a gray jacket converses with an older man in a green shirt and glasses while sitting on a restaurant patio.	\\
2. A blond pony-tailed woman and a gray-haired man converse while seated at a restaurant's outdoor area.	\\
3. A woman with blond hair is sitting at a table and talking to a man with glasses.	\\
4. A woman discusses something with an older man at a table outside a restaurant.	\\
\midrule
A woman doing work.	 \hfill $\Rightarrow$\textsc{\textbf{Neutral}}\\
\end{tabular}
\end{footnotesize}
\end{center}

\section{Models}\label{models}
We apply several neural models from the entailment literature to our data. We also present a model designed to handle multiple premises, as this is unique to our dataset. 

\paragraph{LSTM}\label{lstm-model}
In our experiments, we found that the conditional LSTM \cite{Hochreiter1997} model of \newcite{rocktaschel2016reasoning} outperformed a Siamese LSTM network (e.g. \newcite{Bowman2015}), so we report results using the conditional LSTM. This model consists of two LSTMs that process the hypothesis conditioned on the premise. The first LSTM reads the premise. Its final cell state is used to initialize the cell state of the second LSTM, which reads the hypothesis. The resulting premise vector and hypothesis vector are concatenated and passed through a hidden layer and a softmax prediction layer. When handling four MPE premise sentences, we concatenate them into a single sequence (in the order of the caption IDs) that we pass to the first LSTM. When we only have a single premise sentence, we simply pass it to the first LSTM.

\paragraph{Word-to-word attention}
Neural attention models have shown a lot of success on SNLI. We evaluate the word-to-word attention model of \newcite{rocktaschel2016reasoning}.\footnote{Our experiments use a reimplementation of their model \scriptsize{\url{https://github.com/junfenglx/reasoning_attention}}} 
This model learns a soft alignment of words in the premise and hypothesis. One LSTM reads the premise and produces an output vector after each word. A second LSTM, initialized by the final cell state of the first, reads the hypothesis one word at a time. For each word $w_t$ in the hypothesis, the model produces attention weights $\alpha_t$ over the premise output vectors. The final sentence pair representation is a nonlinear combination of the attention-weighted representation of the premise and the final output vector from the hypothesis LSTM. This final sentence pair representation is passed through a softmax layer to compute the cross-entropy loss. Again, when training on MPE, we concatenate the premise sentences into a single sequence as input to the premise LSTM.

\paragraph{Premise-wise sum of experts (SE)}
The previous models all assume that the premise is a single sentence, so in order to apply them naively to our dataset, we have to concatenate the four premises. 
To capture what distinguishes our task from standard entailment, we also consider a premise-wise sum of experts (SE) model that makes four independent decisions for each premise paired with the hypothesis. This model can adjust how it processes each premise based on the relative predictions of the other premises.

We apply the conditional LSTM repeatedly to read each premise and the hypothesis, producing four premise vectors $p_1$ ... $p_4$ and four hypothesis vectors $h_1$ ... $h_4$ (conditioned on each premise). Each premise vector $p_i$ is concatenated with its hypothesis vector $h_i$ and passed through a feedforward layer to produce logit prediction $l_i$. We sum $l_1$ ... $l_4$ to obtain the final prediction, which we use to compute the cross-entropy loss. 

When training on SNLI, we apply the conditional LSTM only once to read the premise and hypothesis and produce $p_1$ and $h_1$. We pass the concatenation of $p_1$ and $h_1$ through the feedforward layer to produce $l_1$, which we use to compute the cross-entropy loss. 

\begin{table}
\begin{center}
\begin{small}
\begin{tabular}{llrrr}
	\toprule
	Training & Class & LSTM & SE & Attention \\	
	\midrule
	SNLI only & & 52.6 & \textbf{55.9} & 55.0 \\
    		& E & 85.8 & 71.5 & 81.7 \\
            & N & 8.4 & 21.6 & 16.4 \\
            & C & 55.7 & 62.0 & 54.5 \\
    \midrule
	MPE only & & 53.5 & \textbf{56.3} & 53.9 \\
    		& E & 63.1 & 61.3 & 48.3 \\
            & N & 39.2 & 30.2 & 30.6 \\
            & C & 53.5 & 66.5 & 71.2 \\
    \midrule
	SNLI+MPE & & 60.4 & 60.0 & \textbf{64.0} \\
    		& E & 65.1 & 65.4 & 75.9 \\
            & N & 40.9 & 42.7 & 32.8 \\
            & C & 67.2 & 65.1 & 71.5 \\
	\bottomrule
\end{tabular}
\end{small}
\caption{Entailment accuracy on MPE (test).  SE  is best when training only on SNLI or  MPE. Attention is  best when training on SNLI+MPE.}
\vspace{-10pt}
\label{tab:results}
\end{center}
\end{table}

\section{Training Details}
For the LSTM and SE models, we use 300d GloVe vectors \cite{pennington2014glove} trained on 840B tokens as the input. The attention model uses word2vec vectors \cite{mikolov2013distributed} (replacing with GloVe had almost no effect on performance). We use the Adam optimizer \cite{adamKingma} with the default configuration. We train each model for 10 epochs based on convergence on dev. For joint SNLI+MPE training, we use the same parameters and pretrain for 10 epochs on SNLI, then train for 10 epochs on MPE. This was the best joint training approach we found.

When training on SNLI, we use the best parameters reported for the word-to-word attention model.\footnote{Dropout: 0.8, learning rate: 0.001, vector dim: 100, batch size: 32} When training on MPE only, we set dropout, learning rate, and LSTM dimensionality as the result of a grid search on dev.\footnote{LSTM: dropout: 0.8, vector dim: 75. SE: dropout: 0.8,  vector dim: 100. Attention: dropout: 0.6, vector dim: 100. Learning rate: 0.001 for all models}

\section{Experimental Results}

\subsection{Overall Performance}
Table \ref{tab:results} contains the test accuracies of the models from Section~\ref{models}: LSTM, sum of experts (SE), and word-to-word attention under three training regimes: SNLI only, MPE only, and SNLI+MPE. 

We train only on SNLI to see whether models can generalize from one entailment task to the other. Interestingly, the attention model's accuracy on MPE is higher after training only on SNLI than training on MPE, perhaps because it requires much more data to learn reasonable attention weighting parameters. 

When training on SNLI or MPE alone, the best model is SE, the only model that handles the four premises. It is not surprising that the LSTM model performs poorly, as it is forced to reduce a very long sequence of words to a single vector. The LSTM performs on par with SE when training on SNLI+MPE, but our analysis (Section~\ref{sec:discussion}) shows that their errors  are quite different.

The attention model trained on SNLI+MPE has the highest accuracy overall. We reason that pretraining on SNLI is necessary to learn reasonable parameters for the attention weights before training on MPE, a smaller dataset where word-to-word inferences may be less obvious. When trained only on MPE, the attention model performs much worse than SE, with particularly low accuracy on entailing items.

We implemented a model that adds attention to the SE model, but it overfit on SNLI and could not match other models' accuracy, reaching only about 58\% on dev compared to 59-63\%. Future work will investigate other approaches to combining the benefits of the SE and attention models.

\begin{table}
\centering
\begin{small}
\begin{tabular}{lrrrrr}
	\toprule
    \multicolumn{6}{c}{\textbf{Accuracy on SPE-MPE agreement subsets}} \\
    \# pairs agree & 0 & 1 & 2 & 3 & 4	 \\
    \% of data & 21.8 & 26.9 & 16.7 & 24.8 & 9.8 \\
    \midrule
    LSTM & 57.3 & 57.6 & 60.5 & 67.1 & 63.3 \\
    SE & 59.6 & \textbf{58.0} & \textbf{63.3} & 62.9 & 66.3 \\
    Attention & \textbf{65.6} & 57.6 & 62.9 & \textbf{68.3} & \textbf{70.4} \\
    \bottomrule
\end{tabular}
\end{small}
\caption{Accuracy for each model (trained on SNLI+MPE) on the dev data subsets that have 0--4 SPE labels that match the MPE label (Table~\ref{tab:examples_difficulty}).}
\vspace{-10pt}
\label{tab:pairs_analysis}
\end{table}

\subsection{Performance by Pair Agreement}
To get a better understanding of how our task differs from standard entailment, we analyze how performance is affected by the number of premises whose SPE label agrees with the MPE label. Table~\ref{tab:pairs_analysis} shows the accuracy of each SNLI+MPE-trained model on the dev data grouped by SPE-MPE label agreement (as in Table~\ref{tab:examples_difficulty}).

The attention model has the highest accuracy on three of five categories, including the most difficult category where none of the SPE labels match the MPE label. SE has the highest accuracy in the remaining two categories. The attention model demonstrates large gains in the easiest categories, perhaps because there is less advantage to aggregating individual premise predictions (as SE does) and more cases where attention weighting of individual words is useful. On the other hand, the attention model also does well on the most difficult category, indicating that it may be able to partially aggregate premises by increasing attention weights on phrases from multiple sentences. Attention and SE exhibit complementary strengths that we hope to combine in future work.



\subsection{Performance by Semantic Phenomenon}
Table \ref{tab:errors} shows the performance of the three SNLI+MPE-trained models over semantic phenomena, based on the 100 annotated dev items (see Section~\ref{sec:discussion} and Table \ref{tab:info_type_pairs}). It may not be informative to analyze performance on smaller classes (e.g. phrase equivalence and phrase hypernymy), but we can still look at other noticeable differences between models.

Although the attention model outperformed both LSTM and SE models in overall accuracy, it is not the best in every category. Both SE and attention have access to the same information, but the attention model does better on items that contain relationships like hypernyms and synonyms for both words and short phrases. The SE model is best at mutual exclusion, compatibility, and world knowledge categories, e.g. knowing that a man who is \textit{resting} is not \textit{kayaking}, and a \textit{bride} is not also a \textit{cheerleader}. In cases that require analysis of mutually exclusive or compatible events, a model like SE has an advantage since it can reinforce its weighted combination prediction by examining each premise separately.

\begin{table}
\centering
\begin{small}
\begin{tabular}{lrrrr}
	\toprule
    &  \multicolumn{3}{c}{Accuracy}  & \\
    Phenomenon & LSTM & SE  & Att & \# \\
    \midrule
    Word equivalence & 50.0 & 56.2 & \textbf{68.8} & 16\\
    Word hypernymy & \textbf{52.6} & 47.4 & \textbf{52.6}& 19 \\
    Phrase equivalence & 57.1 & 57.1 & \textbf{85.7} & 7 \\
    Phrase hypernymy & 50.0 & 50.0 & \textbf{62.5} & 8 \\
    Mutual exclusion & 68.0 & \textbf{72.0} & 60.0 & 25 \\
    Compatibility & 50.0 & \textbf{61.1} & 50.0 & 18 \\
    World knowledge & 57.1 &\textbf{62.9} & 45.7 & 35 \\
    \bottomrule
\end{tabular}
\end{small}
\caption{Accuracy for each semantic phenomenon on 100 dev items. While attention was the best model overall, it does not have the highest accuracy for all phenomena.}
\vspace{-10pt}
\label{tab:errors}
\end{table}

\section{Conclusion}

We presented a novel textual entailment task that involves inference over longer premise texts and aggregation of information from multiple independent premise sentences. This task is an important step towards a system that can create a coherent scene representation from longer texts, such as multiple independent reports. We introduced a dataset for this task ({\scriptsize \href{http://nlp.cs.illinois.edu/HockenmaierGroup/data.html}{\nolinkurl{http://nlp.cs.illinois.edu/HockenmaierGroup/data.html}}}) which presents a more challenging, realistic entailment problem and cannot be solved by majority voting or related heuristics. We presented the results of several strong neural entailment baselines on this dataset, including one model that aggregates information from the predictions of separate premise sentences. Future work will investigate aggregating information at earlier stages to address the cases that require explicit reasoning about the interaction of multiple premises. 

\section*{Acknowledgments}
This project is  supported by NSF grants 1053856, 1205627, 1405883 and 1563727, a Google Research Award, and Contract W911NF-15-1-0461 with the US Defense Advanced Research Projects Agency (DARPA) and the Army Research Office (ARO). Approved for Public Release, Distribution Unlimited. The views expressed are those of the authors and do not reflect the official policy or position of the Department of Defense or the U.S. Government.

\bibliography{ijcnlp2017}
\bibliographystyle{ijcnlp2017}

\end{document}